\documentclass{article}

\PassOptionsToPackage{numbers, compress}{natbib}
\usepackage[preprint]{neurips_2020}

\usepackage[utf8]{inputenc} % allow utf-8 input
\usepackage[T1]{fontenc}    % use 8-bit T1 fonts
\usepackage{hyperref}       % hyperlinks
\usepackage{url}            % simple URL typesetting
\usepackage{booktabs}       % professional-quality tables
\usepackage{amsfonts}       % blackboard math symbols
\usepackage{nicefrac}       % compact symbols for 1/2, etc.
\usepackage{microtype}      % microtypography
\usepackage{graphicx}
\usepackage{color}
\usepackage{bm}
\usepackage{algorithm}
\usepackage{algpseudocode}
\usepackage{amsmath}
\usepackage{ulem} 

\title{Vision at A Glance: Interplay between Fine and Coarse Information Processing Pathways}

\author{%
  Zilong Ji\thanks{Equal contribution} \\
  State Key Laboratory of Cognitive Neuroscience \& Learning, Beijing Normal University, China.\\
  \texttt{jizilong@mail.bnu.edu.cn} \\
  \And
  Xiaolong Zou\footnotemark[1], Tiejun Huang, Si Wu \thanks{Corresponding author}\\
  School of Electronics Engineering \& Computer Science, Peking University, Beijing, China.\\
  \texttt{xiaolz, tjhuang, siwu@pku.edu.cn} \\
}

\begin{document}

\maketitle

\begin{abstract}
Object recognition is often viewed as a feedforward, bottom-up process in machine learning, but in real neural systems, object recognition is a complicated process which involves the interplay between two signal pathways. One is the parvocellular pathway (P-pathway), which is slow and extracts fine features of objects; the other is the magnocellular pathway (M-pathway), which is fast and extracts coarse features of objects. It has been suggested that the interplay between the two pathways endows the neural system with the capacity of processing visual information rapidly, adaptively, and robustly. However, the underlying computational mechanisms remain largely unknown. In this study, we build a computational model to elucidate the computational advantages associated with the interactions between two pathways. Our model consists of two convolution neural networks: one mimics the P-pathway, referred to as FineNet, which is deep, has small-size kernels, and receives detailed visual inputs; the other mimics the M-pathway, referred to as CoarseNet, which is shallow, has large-size kernels, and receives low-pass filtered or binarized visual inputs. The two pathways interact with each other via a Restricted Boltzmann Machine. We find that: 1) FineNet can teach CoarseNet through imitation and improve its performance considerably; 2) CoarseNet can improve the noise robustness of FineNet through association; 3) the output of CoarseNet can serve as a cognitive bias to improve the performance of FineNet. We hope that this study will provide insight into understanding visual information processing and inspire the development of new object recognition architectures.
\end{abstract}

\section{Introduction}
Imagine you are driving a car on a highway and suddenly an object appears in your visual field, crossing the road. Your initial reaction is to slam on the brakes even before recognizing the object. This highlights a core difference between human vision and current machine learning strategies for object recognition. In machine learning, object recognition is often viewed as a feedforward, bottom up process, where image features are extracted from local to global in a hierarchical manner; whereas in human vision, we can capture the gist of an image at a glance without processing the details of it, a crucial ability for animals to survive in competitive natural environments. This strategic difference has been demonstrated by a large volume of experimental data. For example, Sugase et al. found that neurons in the inferior temporal cortex (IT) of macaque monkeys convey the global information of an object much faster than the fine information of it~\citep{Sugase1999Nature}; FMRI and MEG studies on humans showed that the activation of orbitofrontal cortex (OFC) precedes that of the temporal cortex when a blurred object was shown to the subject~\citep{Bar2006PNAS}.

\begin{figure}[tbp]
  \begin{center}
  \centerline{\includegraphics[width=0.7\linewidth]{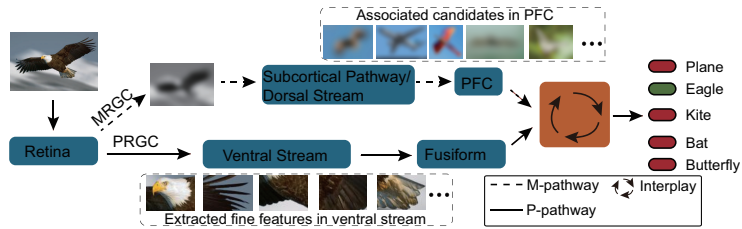}}
  \caption{Two signal pathways for visual information processing in the brain. The P-pathway starts from midget retina ganglion cells (MRGCs) and goes through the ventral stream to Fusiform and other cortical regions. The M-pathway starts from parasol retina ganglion cells (PRGCs) and goes through superior colliculus (SC) or the dorsal stream to prefrontal cortex (PFC) and other cortical
  regions. The P-pathway receives detailed visual inputs and extracts fine features of images. The M-pathway receives low-pass filtered visual inputs and extracts coarse features of images. The two pathways are associated in higher cortical areas to 
  accomplish visual recognition.
  For instance, the M-pathway may generate candidate objects based on the coarse information of an image, which serves as a cognitive bias guiding the fine recognition of the image mediated by the P-pathway.}
  \label{fig-demo}
  \end{center}
\end{figure}

Indeed, the Reverse Hierarchy Theory 
for visual perception has proposed that although the representation of image features in the ventral pathway goes from local to global, our perception of an image goes inversely from global to local~\citep{Hochstein2002Neuron}.
How does this happen in the brain? Experimental studies show that there are two separable signal pathways for visual information processing in the brain
(see Fig.\ref{fig-demo}). One is called the parvocellular pathway (P-pathway), which starts from midget retina ganglion cells (MRGCs), projects to layers 3-6 in the lateral geniculate nucleus (LGN), and then 
primarily  goes downstream along the ventral pathway. The other is called the magnocellular pathway (M-pathway), which starts from parasol retina ganglion cells (PRGCs), projects to layers 1-2 of LGN or the superior colliculus (SC), and then goes downstream to higher cortical areas. The two pathways have different neural response characteristics and complementary computational roles. Specifically, the P-pathway is sensitive to color and responds primarily to visual inputs of high spatial frequency; whereas the M-pathway is color blind and responds primarily to visual inputs of low spatial frequency~\citep{Derrington1984JP}. It has been suggested that the M-pathway serves as a short-cut to extract global information of images rapidly; while the P-pathway extracts fine features of images slowly; the interplay between two pathways endows the neural system with the capacity of processing visual information rapidly, adaptively, and robustly~\citep{Bar2003JCN,Wang2020bioRxiv, Bullier2001BRR}.

Although the P- and M- pathways are well known in the field, exactly how their interplay facilitates object recognition remains poorly understood. 
Conventionally, machine learning learns from neuroscience to develop new models, e.g., convolution neural networks (CNNs) constructed by learning from the hierarchical, feedforward architecture of the visual pathway~\citep{Fukushima1980BC, Lecun1998IEEE} and new machine learning models for scene recognition inspired by the central and peripheral vision~\cite{Wang2017JOV, Wu2018CISS}.
Conversely, recent state-of-the-art machine learning models, such as deep CNNs, were applied to interpret neural data and obtained encouraging results. For example, by using deep CNNs to model the ventral pathway, the response properties of neurons and their computational roles were well interpreted~\citep{Yamins2013nips, Kriegeskorte2015AV}.
In this study, based on the biological relevance and expressive power of CNNs, we build up a two-pathway model to elucidate the computational advantages associated with the interplay between P- and M- pathways. The outcome of this study will naturally give us insight into developing new object recognition architectures.

\begin{figure}[tbp]
  \begin{center}
  \centerline{\includegraphics[width=0.7\linewidth]{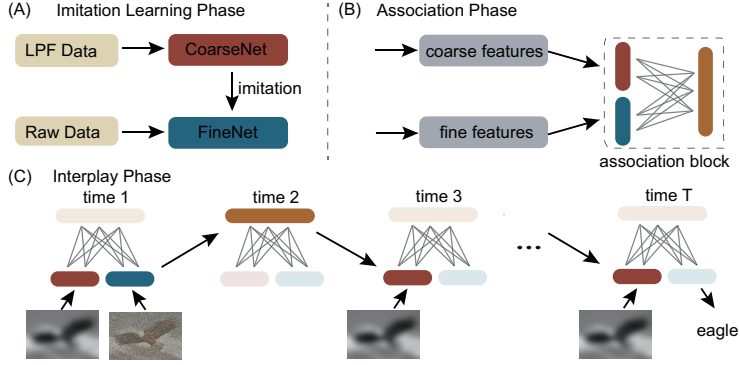}}
  \caption{The two-pathway model consisting of two CNNs (FineNet and CoarseNet) and an RBM. (A) The imitation learning phase. CoarseNet learns from FineNet through imitation. (B) The association phase. FineNet and CoarseNet are associated with each other through the RBM.
  (C) The interplay phase in which CoarseNet improves the noise robustness of FineNet. The example illustrates how the result of CoarseNet helps FineNet recognize a 
  noise corrupted image of eagle via iterating the state of the RBM. LPF: low-pass filtered.} 
  \label{fig-model}
  \end{center}
\end{figure}

Our model consists of three parts, two parallel CNNs and an associative memory network that integrates them (Fig.~\ref{fig-model}). One CNN mimics the P-pathway, which is relatively deep, has small-size kernels, and receives detailed visual inputs. It aims to extract fine features of images, referred to as FineNet hereafter. The other CNN mimics the M-pathway, which is relatively shallow, has large-size kernels, and receives low-pass filtered or binarized visual inputs. It aims to extract coarse features of images, referred to as CoarseNet hereafter. The two CNNs are associated with each other via a Restrict Boltzmann Machine (RBM). Based on this model, we unveil a number of appealing properties associated with the interplay between two pathways. First, since CoarseNet is shallow and unable to learn an object recognition task well, FineNet can teach CoarseNet through imitation to improve its performance considerably. Second, since CoarseNet has large convolution kernels and receives coarse inputs, its performance is robust to noise corruptions, which in return can enhance noise robustness of FineNet via association. Third, since CoarseNet is faster (mimicking the M-pathway), 
its output can serve as a cognitive bias to leverage the performance of FineNet. 

\section{The Two-pathway Model}
The structure of our two-pathway model, together with its learning and functioning processes, are illustrated in Fig.~\ref{fig-model}. 
Since the detailed structures and functions of the P- and M- pathways are far from clear, we adopt two CNNs of different complexities to capture their essential computational properties
(CNNs have been shown to be effective for modeling visual information processing~\citep{Yamins2013nips, Kriegeskorte2015AV}),
with a focus on their different characteristics in feature extraction. Specifically,   
FineNet is deeper than CoarseNet, reflecting that the P-pathway goes through more feature analyzing relays
(V1-V2-V4-IT in the primate brain) than the M-pathway. 
It also has smaller convolution kernels and receives much more detailed visual inputs, i.e., raw visual images with RGB channels, reflecting that PRGCs are color selective and have much smaller receptive fields than MRGCs
anatomically. 
Due to the large receptive field sizes of MRGCs and the electrical couplings between them (which leads to long-range coherent activity in the retina that is believed to be crucial for global object perception~\citep{Roy2017PNAS}), we set CoarseNet to have relatively large-size  
convolution kernels and process low-pass filtered visual inputs.
Moreover, since MRGCs are color blind, we consider that CoarseNet processes grayed images.
Biologically, information processing along the M-pathway is much faster than that along the P-pathway, but this is not reflected in our model, as CNNs do not include the dynamics of neurons. 
Experimental data indicates that multiple brain regions are likely
involved in the association between the two pathways, such as the medial temporal lobe, hippocampus~\citep{Eichenbaum2000, Ranganath2004}, and the parahippocampal cortex~\citep{Aminoff2007}. Here, we simplify this as an associative memory process mediated by an RBM, whereby the outputs of two CNNs can interact with each other.     

\subsection{The imitation learning phase}
Given an input image $\bm{x}$,
denote the output of FineNet to be $\bm{p}^F(\bm{x})=\bm{f}^F\left(\bm{g}^F\left(\bm{x},\bm{\theta}^F\right),\bm{w}^F\right)$, where $\bm{g}^F(\cdot,\bm{\theta}^F)$ represents
the mapping function from the input to the penultimate layer, $\bm{f}^F(\cdot,\bm{w}^F)$ the readout function, and
$\{\bm{\theta}^F,\bm{w}^F\}$ the trainable parameters. $\bm{p}^F(\bm{x})$ is a K-dimensional vector, 
with $K$ the number of image classes. 
Similarly, the output of CoarseNet is written as  
$\bm{p}^C(\hat{\bm{x}})=\bm{f}^C\left(\bm{g}^C\left(\hat{\bm{x}},\bm{\theta}^C\right),\bm{w}^C\right)$, where $\hat{\bm{x}}$ represents the coarse input with respect to the image $\bm{x}$.
$\bm{g}^C(\cdot,\bm{\theta}^C)$ and $\bm{f}^C(\cdot,\bm{w}^C)$ represent, respectively,
the mapping functions from the input to the penultimate layer and 
the readout function, and
$\{\bm{\theta}^C,\bm{w}^C\}$ the trainable parameters. 
The coarse input to CoarseNet is obtained by either low-pass filtering
a grayed image using a 2D Gaussian filter or binarizing an image (see examples in Fig.~\ref{fig-data}A).

First, to get a model of the P-pathway,
we optimize FineNet through minimizing a cross-entropy loss, which is written as
\begin{equation}
    L_F=-\frac{1}{N}\sum^N_{i=1}\sum^K_{j=1}y_{i,j}
    \ln{p^F_j(\bm{x}_i)},
\end{equation}
where $p^F_j$ is the $j$th element of $\bm{p}^F$, i.e., the likelihood of 
the $j$th class, and $y_{i,j}$ is the $j$th element of the
one-hot label $\bm{y}_i$ for the image $\bm{x}_i$, which
is $1$ for the correct class and $0$ otherwise. 
The summation runs over all images $N$ and all classes $K$.

Second, to get a model of the M-pathway,
we optimize CoarseNet via imitation learning from the optimized FineNet, and the corresponding loss function is given by 
\begin{equation}
    L_{C}=\frac{1}{N}\sum^N_{i=1}
    \left[-\alpha
    \sum^K_{j=1}y_{i,j}\ln p^C_j(\hat{\bm{x}_i})+\frac{1-\alpha}{2}\|
    \bm{g}^C(\hat{\bm{x}}_i)-\bm{g}^F(\bm{x}_i)\|^2
    \right],
    \label{eq-imitation}
\end{equation}
where
$\bm{g}^F(\bm{x}_i)$ is the imitation target for feature representations, i.e., the neural activity at the penultimate layer of FineNet. The symbol $\|\cdot \|$ denotes $L_{2}$ normal, and $\alpha$ is a hyper-parameter controlling the balance between the cross-entropy and imitation losses. Through imitating the features representations of FineNet,
CoarseNet learns to classify images based on the coarse inputs. 

\subsection{The association and interplay phases}
\label{asso-interlay}
After training two networks, we perform association learning to establish their correlation via an RBM.
The RBM is a simplified version of the Boltzmann Machine (BM), with the latter being an extension of the Hopfield model including stochastic dynamics~\citep{Hinton2006NC}. Both the BM~\citep{Ackley1985CS} and the Hopfield model~\citep{Hopfield1982PNAS} aim to
capture how memory patterns are stored as stationary states of neural circuits via recurrent connections between neurons. An RBM consists of a visible and a hidden layers with no within-layer connections. 

It is unclear yet how two visual 
pathways interact with each other biologically, which is complex and task-dependent. Hence we consider that the associated information via the RBM is slightly different for the two tasks investigated in the present study.
Specifically, for the task of using the output of CoarseNet to improve the robustness of FineNet to noise, referred to as the robustness task (see Sec.~\ref{sec-robustness}), we
concatenate the neural activities in the penultimate layers of CoarseNet and FineNet in response to the same images, i.e, 
$\left\{\bm{g}^C(\hat{\bm{x}}), \bm{g}^F(\bm{x})\right\}$,
as the inputs to the visible layer of the RBM.
For the task of using the output of CoarseNet as a cognitive bias to improve the performance of FineNet, referred to as the cognitive-bias task (see Sec.~\ref{sec-bias}), we concatenate
the neural activity in the penultimate layer of CoarseNet $\bm{g}^C(\hat{\bm{x}})$ and the corresponding context vector $\bm{c}(\hat{\bm{x}})$ as the inputs to the visible layer of the RBM.
Once the inputs to the visible layer (i.e., the data pairs to be stored) 
are specified, we optimize the connections between the visible and hidden layers, such that the data pairs are stored as local minimums of the energy function of the RBM. 
%The details of training the RBM can be found in Supplementary Information (SI.1).

After training, since the data pairs are stored as memory patterns in the RBM, it is expected that when one component of a data pair is presented as a cue, the RBM will retrieve the other component automatically. The retrieved information can then be used for the recognition task.
Denote the updating dynamics of the RBM to be $[\bm{v}_{t+1}^C,\bm{v}_{t+1}^F]=\bm{A}([\bm{v}_{t}^C,\bm{v}_{t}^F])$, with
$\bm{A}(\cdot)$ the mapping function implemented by the RBM.
In the noise robustness task (Sec.~\ref{sec-robustness}), 
we clamp $\bm{v}_t^C=\bm{g}^C(\hat{\bm{x}})$ unchanged over time,
and set the initial state $\bm{v}_{0}^F=\bm{g}^F(\bm{x})$. After iterating the state of the RBM by a number of steps $T$, the output $\bm{v}_{T}^F$ gives the associated feature representation of FineNet.
In the cognitive-bias task (Sec.~\ref{sec-bias}), 
we clamp $\bm{v}_t^C=\bm{g}^C(\hat{\bm{x}})$ unchanged over time, and set the initial state $\bm{v}_{0}^F=0$. After iterating the state of the RBM by a number of steps $T$, the output $\bm{v}_{T}^F$ gives the retrieved context vector $\bm{c}(\hat{\bm{x}})$ for the image $\bm{x}$. This context vector is then combined with the feature representation of FineNet,
denoted as $\bm{g}^F(\bm{x},\bm{c}(\hat{\bm{x}}),\bm{\theta}^F)$, 
to carry out the recognition.
%For more details, see SI.3.

\section{Interplay between Two Pathways}\label{exp}

\subsection{Implementation details}\label{imple-detail}

\begin{figure}[tbp]
  \begin{center}
  \centerline{\includegraphics[width=0.7\linewidth]{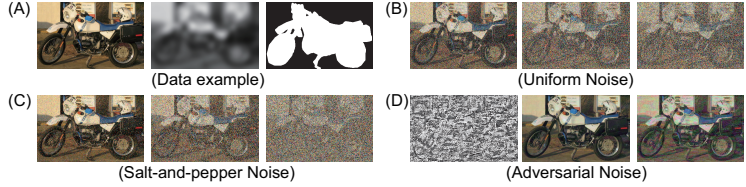}}
  \caption{Examples of visual inputs used in the experiments. (A) Examples of visual inputs used for training FineNet and CoarseNet.
  From left to right, a raw image to FineNet, a low-pass filtered image (LPF data) to CoarseNet, and a binarized image (mask data) to CoarseNet. (B-D) Visual inputs corrupted with different kinds of noise for evaluating the model. (B) Examples with uniform noise sampled from the range of $[-U,U]$. From left to right, the width $U$ is $0.1$, $0.5$, or $0.8$, respectively. (C) Examples with salt-and-pepper noise. From left to right, the proportion of image pixels replaced by white and black ones is  $0.1$, $0.5$, or $0.8$, respectively. (D) Adversarial noise.
  Left: the adversarial noise of the example image in (A)-left, obtained by the Fast Gradient Sign Method~\citep{Goodfellow2015};  Middle and Right: the adversarial examples with noise level of $0.1$ and $0.5$, respectively. }
  \label{fig-data}
  \end{center}
\end{figure}

\begin{figure}[htbp]
  \begin{center}
  \centerline{\includegraphics[width=0.7\linewidth]{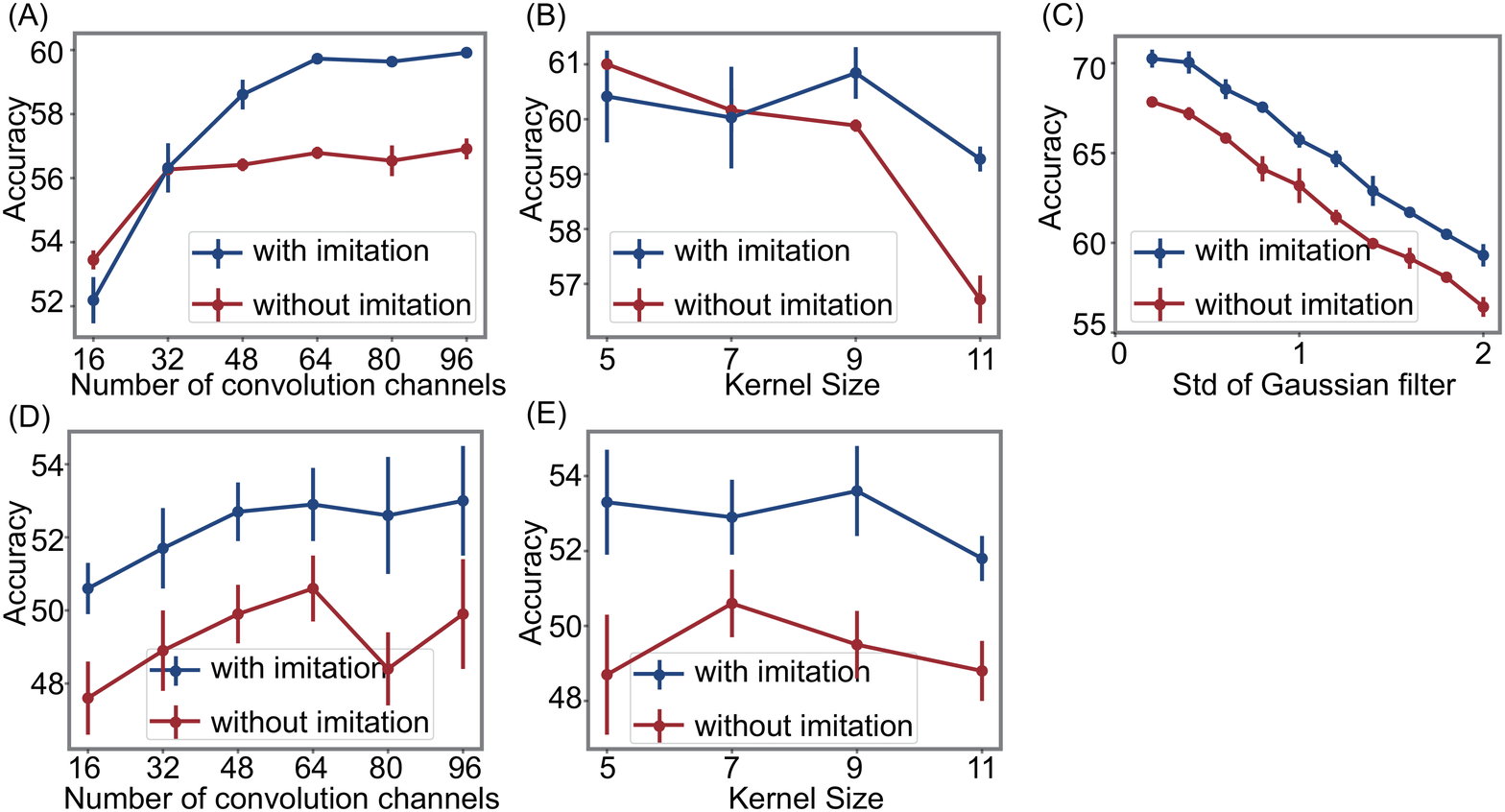}}
  \caption{Imitation learning from FineNet improves the performance of CoarseNet. (A-C): the performances of CoarseNet trained on the low-pass filtered images from CIFAR10 with or without imitation. (A) Performance vs. the number of convolution channels. (B) Performance vs. the size of convolution kernel. (C) Performance vs. STD of the Gaussian filter.  (D-E): the performances of CoarseNet training on the binarized images from Pascalvoc-mask. (D) Performance vs. the number of convolution channels. (E) Performance vs. the size of convolution kernel. Parameters: STD of the Gaussian filter in (A-B) is $2.0$. Other parameters are the same as described in Sec.~\ref{imple-detail}.}
  \label{fig-result1}
  \end{center}
\end{figure}

Based on the proposed model, we carry out simulation experiments to explore the computational properties associated with the interplay between two pathways. Three datasets, CIFAR-10, Pascalvoc-mask and CIFAR-100 are used.
CIFAR-10 is for demonstrating the effect of imitation learning (Fig.~\ref{fig-result1}A-C) and the robustness of the model to noise (Fig.~\ref{fig-result2}). 
Pascalvoc-mask (binarized images) is used as another form of 
coarse
inputs to evaluate the performance of CoarseNet (Fig.~\ref{fig-result1}D-E).
CIFAR-100 is for demonstrating the effect of cognitive bias
(Fig.~\ref{fig-result3}). An example of visual inputs used in the experiments is displayed in Fig.~\ref{fig-data}. 
%For the details of three datasets, see SI.2.

FineNet used in this work consists of three stacked layers, each of which comprises a 128-filter $3\times3$ convolution, followed by a batch normalization, a ReLU nonlinearity, and $2\times2$ max-pooling. CoarseNet has two stacked layers with the same composition as in FineNet, except that it comprises 64-filter $11\times11$ convolution in the first layer and 128-filter $9\times9$ convolution in the second layer. 
The parameter
$\alpha=0.4$ is used when training CoarseNet. Both FineNet and CoarseNet have a fully-connected layer of $1000$ units before the readout layer. Except for normalizing with the channel-wise mean and standard deviation of the whole dataset, no other pre-processing strategies are adopted. 
During the training of FineNet and CoarseNet, the total number of epochs is $150$. SGD with momentum $0.9$, batch size $64$, and an initial learning rate $0.1$ is used. The learning rate is multiplied with $0.1$ after $100$ and $125$ epochs. During the training of the RBM, the total number of epochs is $2000$. We also use SGD to optimize the RBM with an initial learning rate of $0.1$, which is multiplied with $0.1$ after $500$ and $1000$ epochs. The visible and hidden layers of the RBM have $2000$ units and $400$ units, respectively.

\subsection{CoarseNet learning from FineNet via imitation}

The first computational issue we address is about improving the
performance of CoarseNet. Because it is shallow, has large convolution kernels, and receives coarse inputs, CoarseNet is 
normally unable to learn an object recognition task well. Therefore, we explore whether CoarseNet can imitate FineNet to improve its performance (Eq.\ref{eq-imitation}). The implication of the result is discussed in Sec.~\ref{discussion}.

Fig.~\ref{fig-result1} shows that with imitation, the classification accuracy of CoarseNet is improved considerably, compared to that without imitation over a wide range of parameters. 
Specifically, with respect to the number of convolution kernels in CoarseNet, the improvement is significant when the number of kernels 
is large (Fig.~\ref{fig-result1}A for low-pass filtered inputs;
Fig.~\ref{fig-result1}D for binarized inputs);
with respect to the size of kernels in CoarseNet, 
the improvement is also significant 
(Fig.~\ref{fig-result1}B for low-pass filtered inputs; Fig.~\ref{fig-result1}E for binarized inputs);
with respect to variation of the low-pass filter bandwidth
(quantified by the standard deviation (STD) of the Gaussian filter), 
the performance is consistently improved
(Fig.~\ref{fig-result1}C). 
The fact that the effect of imitation learning also depends on
the network parameters (Fig.~\ref{fig-result1}A-B) indicates
that in reality there is a trade-off between
having a simple structure for
the M-pathway and the
capability of the M-pathway imitating the P-pathway.

\subsection{Improved robustness via CoarseNet}
\label{sec-robustness}
The second computational issue we address concerns the robustness of our model to noise.
FineNet, as a deep CNN trained for image classification, is known to overly rely on local textures rather than the global shape of objects, and it is sensitive to unseen noise.
Since CoarseNet processes low-pass filtered (or binarized) visual inputs, whereby the local texture information is no longer the main cue supporting the classification, we expect that the performance of CoarseNet is robust to noise corruptions. Furthermore, through association, we expect that the robustness of FineNet is also leveraged. We carry out simulations to test this hypothesis. 

\begin{figure}[htbp]
  \begin{center}
  \centerline{\includegraphics[width=0.7\linewidth]{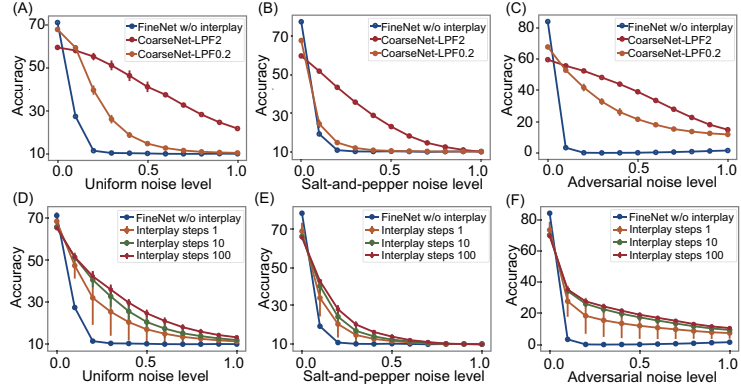}}
  \caption{Robustness of the model to noise. (A-C) CoarseNet is much more robust to noise disruption than FineNet without association. 
  STD of the Gaussian filter is $2.0$ for CoarseNet-LPF2.0 (red line) and $0.2$ for CoarseNet-LPF0.2 (orange line). 
  (D-F) After association, the robustness of FineNet is improved significantly. Interplay steps refers to the number of iteration of the state of the RBM 
  %(see SI.1). 
  (A,D) Performance vs. uniform noise. (B,E) Performance vs. salt-and-pepper noise. (C,F) Performance vs. adversarial noise.  STD of the Gaussian filter is $2.0$ in (D-E). Other parameters are the same as described in Sec.~\ref{imple-detail}.}
  \label{fig-result2}
\end{center} 
\end{figure}

The results are presented in Fig.~\ref{fig-result2}.
We first confirm that CoarseNet is indeed
robust to various forms of noise corruptions,
including uniform noise (Fig.~\ref{fig-result2}A), salt-and-pepper noise (Fig.~\ref{fig-result2}B), and
adversarial noise (Fig.~\ref{fig-result2}C) 
%(for results on binarized inputs, see SI.4 for more details). 
Notably, the robustness of CoarseNet increases when the filter bandwidth
decreases (the red vs. the orange lines). This is understandable, since the filter
bandwidth
controls the spatial precision of visual inputs (the extent of local texture information) to be extracted by CoarseNet. Combining this observation with that in Fig.~\ref{fig-result1} (the accuracy of CoarseNet decreases with the filter bandwidth), it indicates a trade-off between the robustness and accuracy of the network. Remarkably,we observe that through interplay with CoarseNet, the robustness of FineNet is also improved significantly with respect to
uniform noise (Fig.~\ref{fig-result2}D), salt-and-pepper noise (Fig.~\ref{fig-result2}E), and adversarial noise (Fig.~\ref{fig-result2}F). 
The implication of this result is discussed in Sec.~\ref{discussion}.

\subsection{Cognitive bias via the interplay between two networks}
\label{sec-bias}
\begin{figure}[htbp]
  \begin{center}
  \centerline{\includegraphics[width=0.7\linewidth]{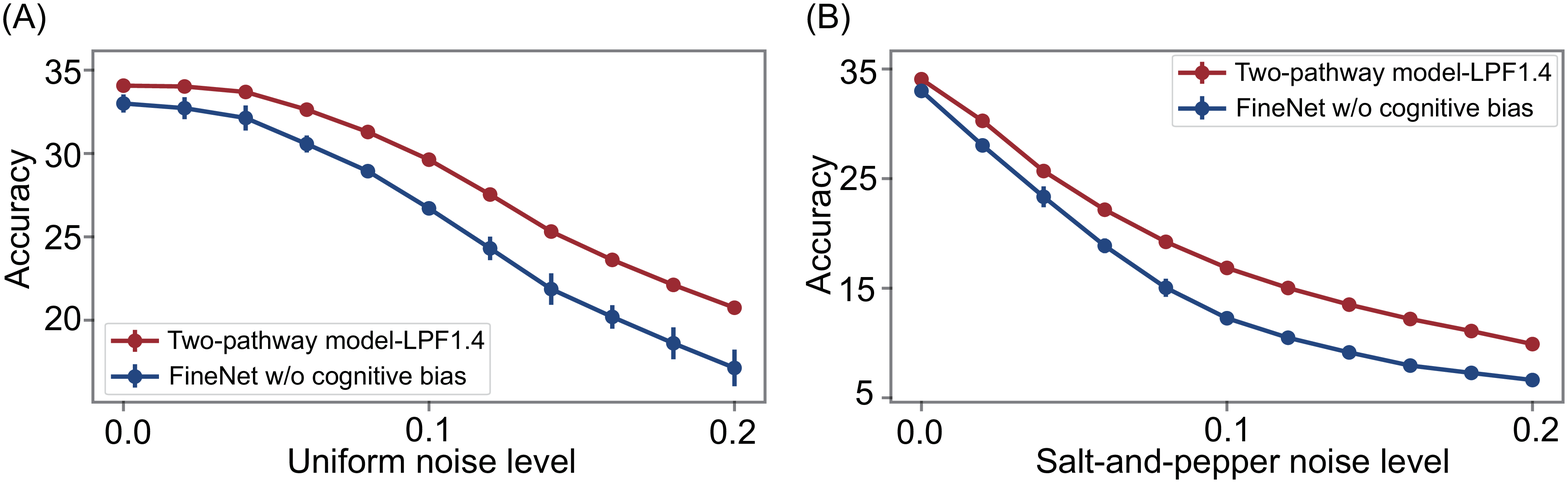}}
  \caption{The output of CoarseNet serving as a cogntive bias which improves the performance of FineNet significantly. The blue line represents the performance of FineNet without a cognitive bias, and the other line
  the performance of FineNet with a cognitive bias generated by CoarseNet. (A)  Performance vs. uniform noise. (B) Performance vs. salt-and-pepper noise. STD of the Gaussian filter is $1.4$.
  Other parameters are the same as described in Sec.~\ref{imple-detail} }
  %Other parameters are the same as described in Sec.~\ref{imple-detail} and SI.3.}
  \label{fig-result3}
\end{center}  
\end{figure}

The third computational issue we address concerns the effect of cognitive bias in visual information process. Cognitive biases, such as the context information, can narrow down object candidates and facilitate recognition significantly. It has been suggested that the output of the M-pathway can serve as a cognitive bias to facilitate the performance of the P-pathway~\citep{Bar2007}. We test this property in our model. Specifically, we consider that the super-class
of an object (e.g., animal) serves as a cognitive bias to help the recognition of the sub-class of the object (e.g., cat).  Each super-class is corresponding to a vector  which can modulate the outputs of FineNet and are called context vectors here.   Images forming
$5$ super-classes and $25$ sub-classes from CIFAR100 are randomly sampled for training
%(for details, see SI.2). 
CoarseNet and FineNet are trained to recognize the super- and sub- class of images, respectively. As described in Sec.~\ref{asso-interlay}, we train the RBM to retrieve 
the context vector conveying the super-class information of an image, when the output of CoarseNet is available. This super-class information is then used to boost the classification of FineNet 
%(for the details, see SI.3). 
The results are presented in Fig.~\ref{fig-result3}, which show that the output of CoarseNet indeed can serve as a cogntive bias to improve the accuracy of FineNet on classifying images based on the sub-class information significantly.  

\section{Conclusion and Discussion}\label{discussion}
In the present study, we have proposed a two-pathway model to mimic the P- and M- signal pathways for visual information processing in the brain. The model is composed of two CNNs, FineNet and CoarseNet, with the former being deeper and having smaller convolution kernels than the latter. The former receives detailed visual inputs and extracts fine features of images, while the latter receives low-passed filtered or binarized visual inputs and extracts coarse
features of images. The two networks interact with each other through an association process mediated by an RBM. We demonstrate that the interplay between the two networks leads to a number of appealing properties. 1) Through imitation from FineNet, the performance of CoarseNet is improved considerably compared to the case of no imitation. 
2) CoarseNet is robust to noise due to its simple structure; and interestingly, through association between two networks, the robustness of FineNet is also improved significantly.
3) the result of CoarseNet can serve as a cognitive bias to leverage the performance of FineNet significantly. 

Although we have used very simple CNNs to mimic the very complicated visual pathways, our model reveals some general advantages
associated with the interplay between two networks of different complexities, which may have some far-reaching implications about our understanding of 
visual information processing in the brain. 

Firstly, an evolutionary drive for the brain to have the M-pathway is for
rapid reaction when facing danger. This speed requirement means that the M-pathway needs to be shallow and process coarse visual inputs in order to save time, but meanwhile it should efficiently generate approximated, if not accurate, recognition of the object, which can serve as a cognitive bias for further improved processing. However, it is a well-known fact that a shallow neural network alone is unable to achieve object recognition well (this has actually motivated the development of deep neural networks). So, how does the brain resolve this dilemma? Here, we argue that the imitation learning strategy proposed in machine learning~\citep{Hinton2015CS} provides a natural solution to this challenge,
that is, the M-pathway can learn from the P-pathway through imitation to improve its performance. Imitation learning has been widely observed in human behaviors, although previous studies mainly focus on imitation between individual persons, without exploring whether it can occurs between neural circuits in the brain. We presume that the brain has resources to implement imitation learning across cortical regions, e.g., the widely observed synchronized oscillations between cortical regions may mediate this~\citep{Buzsaki2004Science}. It will be interesting and exciting to investigate this issue in neurophysiological experiments. 

Secondly, a large volume of comparative studies has shown that human vision is much more robust to noise than machine learning models, with the classical example of adversarial noise~\citep{Goodfellow2015}. Here, our model suggests the potential underlying neural mechanism, that is, the brain exploits two separate pathways of different complexities to process visual inputs at different granularities, such that noise that is harmful to one pathway is not harmful to the other; and through interplay, both pathways eventually become robust to the noise. Previous studies also demonstrated that human vision is robust to binarized images, whereas deep CNNs are not~\citep{Baker2018Plos}. The M-pathway naturally accounts for this phenomenon as demonstrated by CoarseNet in our model.

Thirdly, experimental findings have shown that early activation in the PFC (orbitofrontal cortex in particular) is elicited by the low-pass filtered information from the M-pathway, which serves as a rapid detector or predictor of potential content based on the coarse information of the input (i.e., gist)~\citep{Bar2003JCN}. This rapid predictor may modulate neural activation in the inferior temporal cortex through feedback, and facilitates recognition by biasing the bottom-up process to concentrate on a small set of the most likely object representations. We partly demonstrate this cognitive bias effect using the two-pathway model. This is different from
the previous attentional neural network~\citep{Wang2014NIPS}, which has adopted the label information related to an object as feedback to modulate features, but without specifying how the label information is generated. Our model suggests a more biologically plausible way to implement the cognitive feedback.

Our proposed two-pathway model is also inspirational to machine learning. Deep neural networks (DNNs), which mainly mimic the P-pathway, have proved to be very effective in many applications~\citep{Krizhevsky2012NIPS, He2016CVPR}, yet they still suffer from a lot of shortcomings, including sensitivity to unseen noise. As demonstrated in this study, the two-pathway model provides a potential new architecture to solve this noise sensitivity problem. Conventionally, DNNs focus on extracting local features of images. Recently, there are also efforts aiming to extend DNNs to process ``global information" of images. The methods proposed include, for instance, augmenting training data to have various variations in local features of objects, such that the network is forced to learn information about the global-shape of objects in order to accomplish the recognition task~\citep{Geirhos2019ICLR}; or inducing recurrent connections between neurons in the same layer, so that the network is able to extract texture information over a wide range~\citep{Montobbio2019ARXIV, Spoerer2017FP}. Nevertheless, these methods are very different from ours. Specifically, our model considers having an extra pathway with a coarse structure to extract the global information of images. Furthermore, our model holds advanced cognitive capabilities not shared by a single pathway model. For instance, as partly demonstrated in this work, CoarseNet can quickly capture the gist of an image, which can subsequently serve as a cognitive bias to guide FineNet to perform fine recognition of the image. We expect that the two-pathway model will inspire us to develop new network architectures for implementing more human-like object recognition behaviors. 

% Bibliography
%\newpage
\clearpage

%\section*{Broader Impact}
%In this work, we propose a two-pathway model mimicking the P- and M- pathways for visual information processing in the brain. Based on the model, we unveil a number of appealing computational properties associated with the interplay between two pathways. Our study has potential broad impacts on both brain science and machine learning. First, our model gives new insight into the computational mechanisms of visual information processing, especially, for higher visual cognitive functions which have been rarely studied in previous modelling works. Second, the proposed two-pathway architecture may inspire the development of new object recognition algorithms in machine learning. Ethical issues are not relevant to this theoretical work.

\end{document}